\documentclass[runningheads]{llncs}
\usepackage[T1]{fontenc}
\usepackage{graphicx}
\usepackage{amsfonts}
\usepackage{booktabs}
\usepackage{dsfont}
\usepackage{multirow}
\usepackage{amssymb}
\usepackage{subfigure}
\usepackage{amsmath, amssymb} 
\usepackage{algorithm}
\usepackage{algpseudocode} 
\usepackage{float}
\usepackage{pifont}
\usepackage{multicol}
\usepackage{soul}
\usepackage{nicefrac}       
\usepackage{microtype}      
\usepackage{graphicx}
\usepackage{float}
\usepackage{amsmath}
\usepackage{bm}
\usepackage{makecell}
\usepackage{bbding}

\begin{document}
\title{Point-Supervised Skeleton-Based Human Action Segmentation}
%
%

\author{Hongsong Wang\inst{1} \and
	Yiqin Shen\inst{2} \and
	Pengbo Yan\inst{3} \and
	Jie Gui\inst{2}\textsuperscript{(\Envelope)}\and
	Yang Zhang\inst{4}
}

\authorrunning{H. Wang et al.}
%
\institute{School of Computer Science and Engineering, Southeast University, Nanjing 210096, China \and
	School of Cyber Science and Engineering, Southeast University, Nanjing 210096, China \and
	Southeast University - Monash University Joint Graduate School, SuZhou 215123, China \and
	School of Computer Science and Software Engineering, Shenzhen University, Shenzhen 518060, China\\
	\email{\{hongsongwang,220255055,guijie\}@seu.edu.cn, 18350680668@163.com, yangzhang@szu.edu.cn}}

\maketitle              
\begin{abstract}
Skeleton-based temporal action segmentation is a fundamental yet challenging task, playing a crucial role in enabling intelligent systems to perceive and respond to human activities. While fully-supervised methods achieve satisfactory performance, they require costly frame-level annotations and are sensitive to ambiguous action boundaries. To address these issues, we introduce a point-supervised framework for skeleton-based action segmentation, where only a single frame per action segment is labeled. We leverage multimodal skeleton data, including joint, bone, and motion information, encoded via a pretrained unified model to extract rich feature representations. To generate reliable pseudo-labels, we propose a novel prototype similarity method and integrate it with two existing methods: energy function and constrained K-Medoids clustering. Multimodal pseudo-label integration is proposed to enhance the reliability of the pseudo-label and guide the model training. We establish new benchmarks on PKU-MMD (X-Sub and X-View), MCFS-22, and MCFS-130, and implement baselines for point-supervised skeleton-based human action segmentation. Extensive experiments show that our method achieves competitive performance, even surpassing some fully-supervised methods while significantly reducing annotation effort.

\end{abstract}
\section{Introduction}
Skeleton-based temporal action segmentation~\cite{wang2025foundation} has broad applications in digital humans and robotics, enabling the understanding and segmentation of continuous behaviors. Such capabilities are essential for modeling and generating realistic human motions in digital humans~\cite{wang2026pamd,wang2026temporal,wang2024controllable}, as well as for facilitating human-robot interaction, and adaptive decision-making in dynamic environments.

\begin{figure}[tbp]
	\centering
	\includegraphics[width=0.8\textwidth]{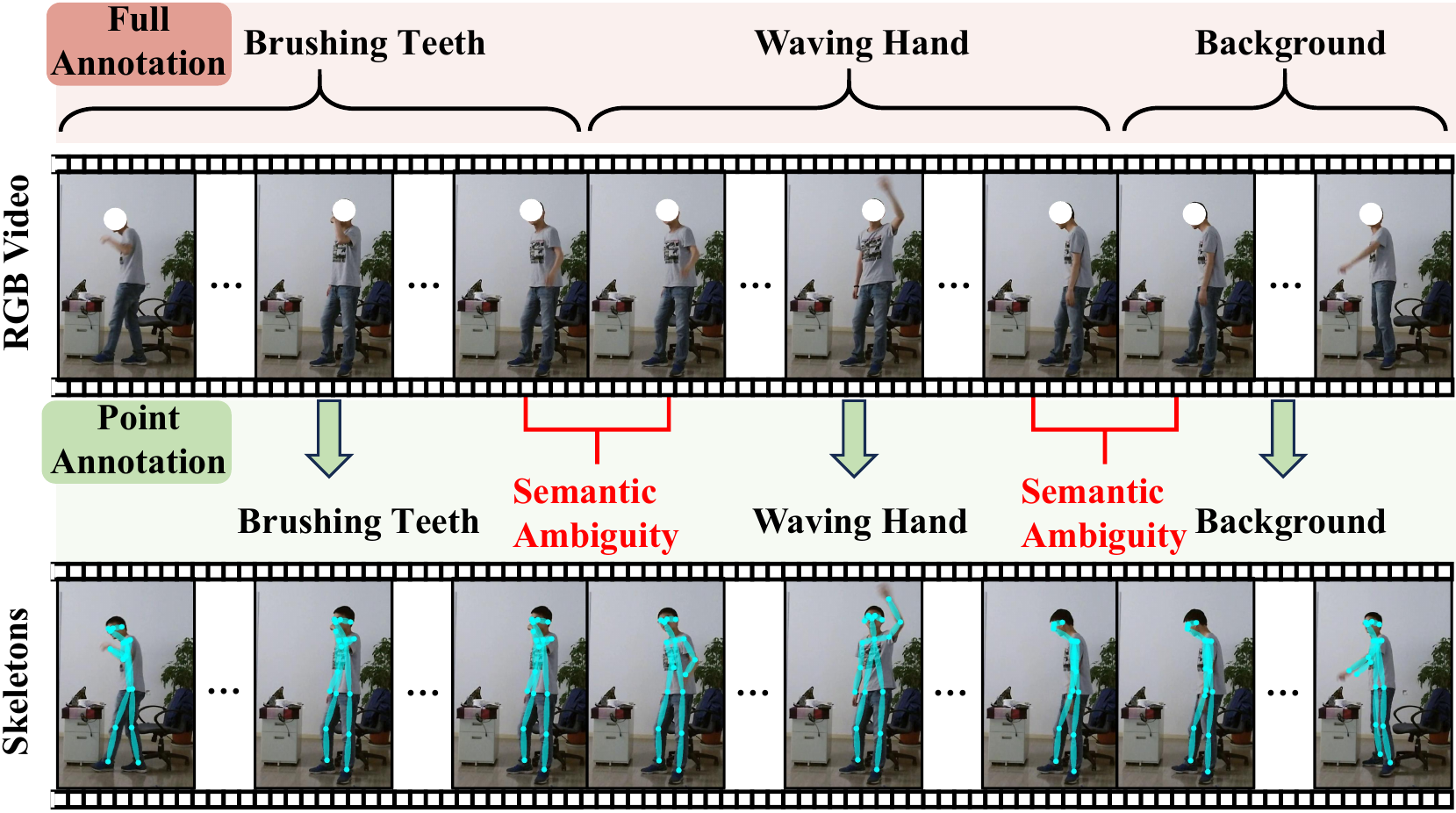} 
	\caption{Comparison between point-supervised annotation and fully-supervised annotation for skeleton-based action segmentation.}
	\label{fig:intro}
\end{figure}

Existing methods primarily focus on spatio-temporal modeling to capture spatial relationships between joints and temporal dependencies across frames to improve frame-level classification features. For spatial modeling, graph convolutional networks or attention mechanisms are commonly employed to model joint correlations. For temporal modeling, temporal convolutional networks or attention mechanisms are used to capture long-term relationships. For instance, LAC \cite{yang2023lac} introduces a linear action composition mechanism in the latent space, which synthesizes and represents new actions by linearly combining skeletal motion features over orthogonal bases. Similarly, Filtjens et al. \cite{filtjens2022skeleton} propose a multi-stage spatial-temporal graph convolutional network that models spatial dependencies among joints with graph convolutions, captures long-term dynamics with extended temporal convolutions. 
However, these methods predominantly rely on fully annotated videos, where the start and end frames of each action must be explicitly labeled, making the annotation process highly time-consuming.

In addition to the high annotation cost, another issue is the ambiguous boundaries between actions for frame-wise annotations. Annotators often struggle to accurately define the transition frames between two neighboring actions. As illustrated in Fig. \ref{fig:intro}, the ending frames of ``brushing teeth” and the starting frames of ``waving hand” can be highly similar, leading to inconsistent segmentation choices among different annotators. 

To address these challenges, we introduce point supervision for skeleton-based action segmentation. This point annotation requires only one annotated point per action segment and background segment. Compared to traditional full supervision, this labeling strategy is not only more efficient but also facilitates communication with annotators. Annotators no longer need to precisely determine action boundaries; instead, they only mark a frame of interest within each segment. This significantly reduces annotation cost and time, providing a practical foundation for skeleton-based action segmentation.

As skeleton data involves complex spatial and temporal structures, we adopt the pretrained unified multi-modal model \cite{sun2023unified} to extract features from three modalities, i.e., joint, bone and motion. Next, pseudo-label generation and integration methods are designed to obtain reliable frame-wise pseudo labels derived from point labels. Finally, the MS-TCN segmentation network \cite{farha2019ms} is employed for training under supervision from the generated pseudo-labels. To summarize, our contributions are as follows:
\begin{itemize}
	\item \textbf{Novel task for skeleton-based action segmentation:}
	We introduce a novel and point-supervised setting for skeleton-based temporal action segmentation. Only a single frame per action segment is labeled, overcoming both issues of costly frame-wise annotation and ambiguous action boundaries.
	
	\item \textbf{Effective pseudo-label generation for action segmentation:} We propose a prototype similarity method and combine it with two existing pseudo-label generation methods to produce multiple pseudo-label sequences. Integrating pseudo-labels generated from different skeleton modalities enhances their robustness, enabling them to effectively guide model training.
	
	\item \textbf{Benchmarks for point-supervised settings:} We establish benchmarks for point-supervised skeleton-based action segmentation on the PKU-MMD (X-sub and X-view), MCFS-22, and MCFS-130 datasets. Our approach achieves competitive performance across multiple evaluation metrics on these four popular benchmarks, even surpassing state-of-the-art fully supervised methods on certain metrics. We provide point annotations for these datasets to facilitate further research in this field.
\end{itemize}

\section{Related Works}

\noindent\textbf{Skeleton-Based Action Segmentation:}
Unified action understanding~\cite{wang2025foundation,weng2025usdrl,wang2025heterogeneous} plays a crucial role in advancing action segmentation. 
Early studies of skeleton-based action segmentation primarily focus on architectural improvements to Spatial-Temporal Graph Convolutional Networks (ST-GCNs)~\cite{yan2018spatial}. Methods such as the Multi-Stage Spatial-Temporal Graph Convolutional Network (MS-GCN)~\cite{filtjens2022skeleton} establish a strong baseline by combining the spatial modeling of ST-GCN with the multi-stage temporal refinement of MS-TCN \cite{farha2019ms}. 
IDT-GCN \cite{li2023involving} introduces distinguished graph convolutions to model complex joint relationships and alleviate feature over-smoothing. Chai et al. \cite{chai2024motion} incorporate motion-aware inputs and multi-scale temporal convolutions to better capture dynamic motion patterns. DeST~\cite{li2023decoupled} uses joint-specific temporal modeling to capture distinctive motion characteristics and prevent feature degradation in long sequences.

Alternative skeletal representations and learning paradigms are also explored to reduce dependency on annotated data. Unsupervised approaches initiate fine-grained motion primitive discovery without labels~\cite{ma2021fine}. Yang et al. \cite{yang2023lac} develop a self-supervised framework using latent action composition for representation learning. Xu et al. \cite{xu2023efficient} address few-shot scenarios through motion interpolation and temporal alignment techniques. Hyder et al. \cite{hyder2024action} propose 2D skeleton heatmaps as skeletal representations. Recently, LaSA~\cite{ji2024language} incorporates linguistic supervision to provide semantic priors for modeling joint relationships and distinguishing actions. Ji et al. \cite{ji2025snippet} adopt a Snippet-aware attention mechanism through fine-grained spatio-temporal modeling.

\noindent\textbf{Point-Supervised Action Localization:}
Point-supervised methods strike a balance between labeling cost and performance in temporal action localization. Moltisanti et al.~\cite{moltisanti2019action} use single timestamp annotations per action. SF-Net~\cite{ma2020sf} expands this by propagating labels to adjacent snippets for pseudo-label training. While cost-effective, early methods suffer from incomplete action coverage. LACP~\cite{lee2021learning} introduces contrastive learning for action completeness, using a greedy algorithm to maximize outer-inner contrast scores. 

To overcome snippet-level limitations, proposal-based methods emerge to better capture temporal structures. Ju et al.~\cite{ju2021divide} developed a two-stage framework that processes video clips sequentially. 
Action Proposal Network (APN)~\cite{yin2023proposal} features constrained k-medoids clustering for pseudo-label generation and fine-grained contrastive loss for boundary refinement. 
Feature enhancement methods such as CRRC-Net~\cite{fu2022compact} utilize class-wise prototypes to achieve more reliable classification, while SQL-Net~\cite{wang2024sql} adopts a two-stage framework that emphasizes semantic consistency and completeness learning.
Recent advances focus on boundary detection and confidence alignment. SMBD~\cite{liu2024stepwise} introduces a multi-grained boundary detector using dual branches for action and scene change detectio. TSP-Net~\cite{xia2024realigning} addresses confidence misalignment through temporal saliency learning and center score adaptation. HR-Pro~\cite{zhang2024hr} implementes hierarchical reliability propagation using memory modules to store and share snippet prototypes across videos. QROT~\cite{liu2025boosting} integrates query reformation and optimal transport to generate high-quality pseudo labels.

\section{Approach}
\begin{figure*}[htbp]
	\centering
	\includegraphics[width=1\textwidth]{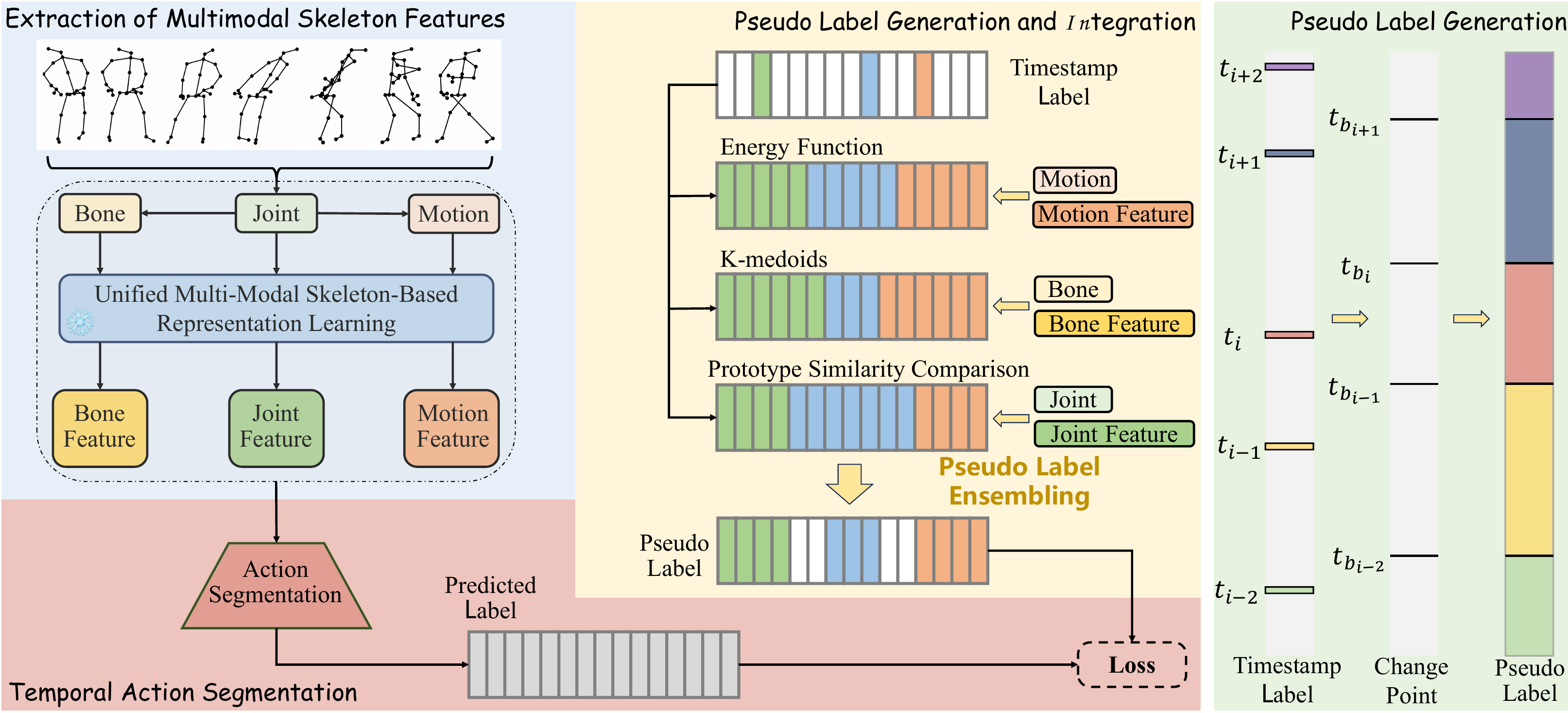} 
	\caption{Overall framework of point-supervised skeleton-based temporal action segmentation via multimodal pseudo-label generation and integration.}
	\label{fig:method}
\end{figure*}

The overall framework of our approach is illustrated in Fig.~\ref{fig:method}.  which includes: (a) extraction of multimodal skeleton features, (b) pseudo-label generation and integration.
The model is optimized end-to-end by minimizing the loss between the predictions and the pseudo-labels, leveraging multimodal characteristics of skeleton data and advantages of point-level supervision. Through modality feature fusion and pseudo-label integration, it effectively improves the accuracy and robustness of skeleton-based temporal action segmentation.

\subsection{Multimodal Feature Extraction}

In skeleton-based action recognition, joint modality $X$ is the original skeleton sequence. Based on $X$, bone $B$ and motion $M$ modalities can be derived to enrich structural and dynamic information~\cite{wang2018beyond}, thereby providing complementary cues for temporal action segmentation.
The bone modality captures structural relations by computing relative positions between adjacent joints, while the motion modality describes temporal dynamics by joint displacements between consecutive frames.

While joint, bone, and motion modalities provide complementary perspectives, they remain low-dimensional and insufficient for direct training. To enhance representation capacity, we adopt UmURL~\cite{sun2023unified} for feature extraction, which is formulated as:
\begin{equation}
	JF = f(X), BF = f(B), MF = f(M),
	\label{equ:umurl}
\end{equation}
where $f(\cdot)$ denotes the UmURL, and $JF, BF, MF \in \mathbb{R}^{T \times D}$ are the extracted joint, bone, and motion features, respectively, $D$ is the feature dimension. The unified backbone for multimodal skeletons reduces computational cost while providing robust features for subsequent pseudo-label generation.

\subsection{Pseudo-Label Generation}
Pseudo-labels provide indirect annotations for unlabeled frames under weak supervision, significantly improving the model’s localization accuracy on unlabeled data. Under point-level supervision, pseudo-label generation can be regarded as identifying an action transition point between two adjacent point annotations: frames preceding the transition point should be assigned to the same category as the left annotation, while frames following the transition point should correspond to the category of the right annotation. 

Furthermore, point-supervised action segmentation can be interpreted as a special clustering problem. For a video containing $N$ action segments, frame features can be clustered into $N$ groups. Since the category label of each point annotation is known, the frames within the corresponding cluster can be assigned labels accordingly. Building on this idea, three pseudo-label generation methods are adopted: the Energy Function~\cite{li2021temporal}, the Constrained K-Medoids clustering~\cite{behrmann2022unified}, and the proposed Prototype Similarity Method. These methods model semantic transitions between frames from the perspectives of distance variation, clustering boundaries, and class prototype similarity, respectively, thereby providing high-quality pseudo-supervision signals for subsequent model training.

\noindent\textbf{Energy Function: }
For each action segment, there exists one annotated frame, and pseudo-label generation can be simplified as finding the action transition point between two adjacent point annotations. Given the video features and annotated points, the energy function aims to locate a point $t_{b_i}$ within the interval $[t_i, t_{i+1}]$, which divides the frames into two clusters such that the sum of distances between frames and their respective cluster centers is minimized:
\begin{align}
	\label{equ:energy1}
	t_{b_i}&=\underset{\hat{t}}{\arg \min } \sum_{t=t_i}^{\hat{t}} d\left(h_t, c_i\right)+\sum_{t=\hat{t}+1}^{t_{i+1}} d\left(h_t, c_{i+1}\right), \\
	c_i&=\frac{1}{\hat{t}-t_i+1} \sum_{t=t_i}^{\hat{t}} h_t, \quad  c_{i+1}=\frac{1}{t_{i+1}-\hat{t}} \sum_{t=\hat{t}+1}^{t_{i+1}} h_t,
\end{align}
where $t_i \leq \hat{t}<t_{i+1}$, $d(.,.)$ denotes the Euclidean distance, $h_t$ represents the feature at time $t$ extracted from the penultimate layer of the action segmentation model, $c_i$ is the average feature between $t_i$ and $\hat{t}$, and $c_{i+1}$ is the average feature between $\hat{t}$ and $t_{i+1}$.

\noindent\textbf{Constrained K-Medoids Clustering: }
The constrained K-Medoids clustering seeks to determine the optimal temporal boundaries of each segment such that the cumulative intra-cluster distance is minimized. Unlike the conventional K-Medoids algorithm, this method explicitly enforces temporal continuity in clustering. It first takes the annotated point frames as the initial clustering centers and calculates the feature distance matrix between all frames. Then, the iterative optimization process begins: the first step is to fix the cluster centers and find the optimal time boundary. Finally, it assigns pseudo-labels to all frames based on the converged time boundary: frames within each time period are assigned the category of the point annotations within that period.

\noindent\textbf{Prototype Similarity Method:}
To fully exploit the point-level annotations, we design a prototype similarity method for pseudo-label generation. Before generating pseudo-labels for each video, all point annotations in the training set are grouped by class, and the mean feature of each class is computed as its prototype representation, denoted as $P \in \mathbb{R}^{C \times D}$. 

The process is as follows: First, initialize the pseudo-label array, and the category label of the first label point will be assigned to the frames before the first label point. For each pair of adjacent point annotation intervals, the method calculates the distance $S_1(t)$ between each frame and the prototype feature of the left annotation point's class, as well as the distance $S_2(t)$ between each frame and the prototype feature of the right annotation point's class, which is formulated as:
\begin{equation}
	\begin{aligned}
		S_1(t) &= \text{dist}(X_t, P_{a_i}), \quad t_i < t < t_{i+1}, \\
		S_2(t) &= \text{dist}(X_t, P_{a_{i+1}}), \quad t_i < t < t_{i+1},
	\end{aligned}
	\label{equ:pro1}
\end{equation}
where $X_t$ denotes the feature at time $t$, $a_i$ and $a_{i+1}$ are the annotated categories of $t_i$ and $t_{i+1}$, respectively, and $P_{a_i}$ and $P_{a_{i+1}}$ are the prototypes of the corresponding categories.

The behavior transition point is determined by finding the frame $t_c$ with the smallest absolute value of the difference between the two distances, and is obtained as:
\begin{equation}
	t_c = \underset{t_i < t < t_{i+1}}{\arg \min} |S_1(t) - S_2(t)|.
	\label{equ:pro2}
\end{equation}
This transition point marks the transitional position where the frame features change from being similar to the left prototype to being similar to the right prototype.

After determining the transition point, this method assigns the category label of the left annotation point to all frames between the left annotation point and the transition point, and the category label of the right annotation point to all frames between the transition point and the right annotation point. This process processes all adjacent annotation point intervals in the video in sequence, and finally assigns the category label of the annotation point to the frame after the last annotation point.

\subsection{Multimodal Pseudo-Label Integration}
Since action boundaries are inherently ambiguous, the temporal intervals near these boundaries are referred as ``ambiguous intervals''. Even for fully supervised settings, different annotators may produce inconsistent labels for these intervals. Due to the semantic uncertainty of frames in ambiguous intervals, different pseudo-label generation methods may assign inconsistent action labels to the same frames. Training with incorrect pseudo-labels in such intervals can lead to error accumulation, deviating the model from the desired outcome. For each video, the aforementioned methods produce three different pseudo-label sequences. Inspired by ensemble learning, the final pseudo-labels are obtained by taking the intersection of these three sequences: if all three methods assign the same label to a frame, it is adopted as the pseudo-label; otherwise, the frame is regarded as ambiguous and left unlabeled. This strategy enhances the reliability of the pseudo-label sequence for guiding model training.

Our approach incorporates paired multimodal data and multimodal features as inputs to different pseudo-label generation methods. The prototype similarity method utilizes joint data and joint features, the constrained K-Medoids method leverages skeleton data and skeleton features, while the energy function is applied to motion data and motion features. Compared to unimodal input, multimodal inputs provide richer perspectives for pseudo-label generation, thereby improving the robustness and stability of the results. 

\section{Experiment}

\subsection{Datasets and Evaluation Metrics}
For skeleton-based action segmentation, our experiments are conducted on the PKU-MMD (X-sub and X-view) \cite{liu2017pku}, MCFS-22 \cite{liu2021temporal}, and MCFS-130 \cite{liu2021temporal} datasets. 

PKU-MMD is a large-scale multimodal dataset for action detection from 3D skeleton sequences. 
The dataset consists of 1009 long untrimmed videos, each containing an average of more than 20 action instances. 
The skeleton modality records the 3D coordinates and confidence scores of 25 joints per frame. PKU-MMD supports two standard evaluation protocols, cross-subject (X-sub) and cross-view  (X-view), designed to assess generalization and viewpoint robustness. 

The Motion-Centered Figure Skating (MCFS) dataset is a domain-specific benchmark for temporal action segmentation, focusing on figure skating. 
Skeletons are extracted and normalized using OpenPose, providing 2D joint coordinates in both 18-point and 25-point formats. Two subsets with different semantic granularities are defined: MCFS-22 and MCFS-130. The MCFS-22 partitions the performances into 22 action categories, emphasizing coarse-grained segmentation and boundary detection. The MCFS-130 refines the annotations into 130 action elements, introducing numerous fine-grained variants with subtle differences.
Both subsets are split into 189 training and 82 testing videos.

We adopt three commonly used metrics in temporal action segmentation: frame-wise accuracy (Acc), edit score (Edit), and segmental F1 score at different temporal IoU thresholds (F1@tIoU). The first metric is frame-based, while the latter two are segment-based. Since temporal segmentation is prone to over-segmentation, where an action is fragmented into multiple discontinuous segments, frame-wise accuracy alone cannot adequately reflect this issue, whereas the Edit and F1 scores are more suitable for evaluating such phenomena.

\noindent\textbf{Frame-wise Accuracy: }
Frame-wise accuracy is the most widely used metric for action segmentation, measuring the percentage of correctly classified frames. However, it does not consider the temporal structure of actions. Even when actions are heavily fragmented, a model may still achieve high accuracy if frame-level predictions are correct.

\noindent\textbf{Edit Score: }
The Edit score quantifies the similarity between two ordered action sequences. It is derived from the edit distance, i.e., the minimum number of insertion, deletion, or substitution operations required to transform the predicted sequence $X$ into the ground-truth sequence $Y$. 
The edit distance $e$ is computed via dynamic programming and then normalized by the maximum length of the two sequences to obtain the final score. As a sequence-level metric, the Edit score evaluates the structural correctness of predicted action sequences without requiring exact frame alignment.


\noindent\textbf{F1 Score: }
The F1 score is computed based on the temporal Intersection over Union (tIoU) between predicted and ground-truth segments. A predicted segment is considered a true positive if its tIoU with a ground-truth segment exceeds a predefined threshold. 
For each ground-truth segment, only the first matched prediction is counted as a true positive, while the rest are treated as false positives. 
Typical thresholds are set to 10\%, 25\%, and 50\%, yielding F1@10, F1@25, and F1@50. Since the F1 score depends on the overlap between predicted and ground-truth segments, it is highly sensitive to over-segmentation and thus provides an effective measure of boundary smoothness.


\subsection{Experimental Setup}
Since this work is the first to introduce the task of point-supervised temporal action segmentation based on skeleton data, existing datasets do not provide corresponding point-level annotations. To address this issue, we construct point-supervised labels from the original frame-level annotations. Specifically, for each annotated action segment, one frame is randomly selected as the point annotation, and its category is assigned as the unique label of the segment, while the labels of the remaining frames are unavailable during training. This label construction strategy is applied consistently across all datasets used in our experiments, including PKU-MMD (X-sub), PKU-MMD (X-view), MCFS-22, and MCFS-130.

We adopt the UmURL~\cite{sun2023unified} pretrained on NTU-60 as the skeleton feature extractor, and the extracted feature dimension is 2048. For the segmentation model, we employ a four-stage MS-TCN~\cite{farha2019ms}, where the first stage contains two parallel convolutional branches with kernel sizes of 3 and 5, respectively, and their outputs are summed before being passed to the next stage. The model is optimized using Adam with a learning rate of 0.0005 and a batch size of 8. The training schedule is set to 120 epochs for PKU-MMD and 150 epochs for MCFS-22 and MCFS-130. During the first 50 epochs, only point annotations are used to obtain a good initialization, after which pseudo-labels generated by the model are progressively incorporated. The loss function hyperparameters are set as $\lambda = 0.15$, $\beta = 0.075$, and $\gamma = 0.15$. For evaluation, we adopt the cross-subject and cross-view splits on the PKU-MMD, while five-fold cross-validation is performed on MCFS-22 and MCFS-130, and the final results are reported as the average performance over the five splits.

\begin{table}[t]
	\centering
	\caption{Results of skeleton-based temporal action segmentation on the PKU-MMD. $\dag$ denotes the method with results reproduced by ours.}
	\resizebox{1\columnwidth}{!}{
		\begin{tabular}{c|l|ccccc|ccccc}
			\toprule
			\multirow{2}[4]{*}{Supervision} & \multicolumn{1}{c|}{Dataset} & \multicolumn{5}{c|}{PKU-MMD (X-sub)}  & \multicolumn{5}{c}{PKU-MMD (X-view)} \\
			\cmidrule{2-12}          & \multicolumn{1}{c|}{Metric} & Acc   & Edit  & F1@10 & F1@25 & F1@50 & Acc   & Edit  & F1@10 & F1@25 & F1@50 \\
			\midrule
			\multirow{6}{1.7cm}{\centering Fully-\\Supervised}  & MS-TCN\cite{farha2019ms} & 65.5  & -     & -     & -     & 46.3  & 58.2  & 56.6  & 58.6  & 53.6  & 39.4  \\
			& MS-GCN \cite{filtjens2022skeleton} & 68.5  & -     & -     & -     & 51.6  & 65.3  & 58.1  & 61.3  & 56.7  & 44.1  \\
			& CTC \cite{xu2023efficient}   & 69.2  & -     & 69.9  & 66.4  & 53.8  & -     & -     & -     & -     & - \\
			& DeST-TCN \cite{li2023decoupled} & 67.6  & 66.3  & 71.7  & 68.0  & 55.5  & 62.4  & 58.2  & 63.2  & 59.2  & 47.6  \\
			& DeST-Former \cite{li2023decoupled} & 70.3  & 69.3  & 74.5  & 71.0  & 58.7  & 67.3  & 64.7  & 69.3  & 65.6  & 52.0  \\
			& LaSA \cite{ji2024language}  & \textbf{73.5} & \textbf{73.4} & \textbf{78.3} & \textbf{74.8} & \textbf{63.6} & \textbf{69.5} & \textbf{67.8} & \textbf{72.9} & \textbf{69.2} & \textbf{57.0} \\
			\midrule
			\multirow{3}{1.7cm}{\centering Point-\\Supervised} & TS-Sup$\,\dag$\cite{li2021temporal} & 61.4  & 63.0  & 65.4  & 58.8  & 41.9  & 65.7  & 67.7  & 70.6  & 64.0  & 45.7  \\
			& TSASPC$\,\dag$\cite{du2022timestamp} & 58.3  & 63.9  & 65.0  & 57.9  & 39.3  & 62.6  & 65.7  & 68.7  & 62.2  & 41.9  \\
			& \textbf{Ours} & \textbf{61.6} & \textbf{67.0} & \textbf{69.6} & \textbf{62.5} & \textbf{44.2} & \textbf{67.1} & \textbf{68.8} & \textbf{73.5} & \textbf{67.2} & \textbf{49.5} \\
			\bottomrule
		\end{tabular}
	}
	\label{tab:PKU}
\end{table}

\begin{table}[t]
	\centering
	\caption{Skeleton-based temporal action segmentation on the MCFS-22 and MCFS-130. }
	\resizebox{1\columnwidth}{!}{
		\begin{tabular}{c|l|ccccc|ccccc}
			\toprule
			\multirow{2}[4]{*}{Supervision} & \multicolumn{1}{c|}{Dataset}  & \multicolumn{5}{c|}{MCFS-22}          & \multicolumn{5}{c}{MCFS-130} \\
			\cmidrule{2-12}          & \multicolumn{1}{c|}{Metric}  & Acc   & Edit  & F1@10 & F1@25 & F1@50 & Acc   & Edit  & F1@10 & F1@25 & F1@50 \\
			\midrule
			\multirow{5}{1.7cm}{\centering Fully-\\Supervised} & MS-TCN \cite{farha2019ms} & 75.6  & 74.2  & 74.3  & 69.7  & 59.5  & 65.7  & 54.5  & 56.4  & 52.2  & 42.5  \\
			& MS-GCN \cite{filtjens2022skeleton} & 75.5  & 72.6  & 75.7  & 70.5  & 57.9  & 64.9  & 52.6  & 52.4  & 48.8  & 39.1  \\
			& DeST-TCN \cite{li2023decoupled} & 78.7  & 82.3  & 86.6  & 83.5  & 73.2  & 70.5  & 73.8  & 74.0  & 70.7  & 61.8  \\
			& DeST-Former \cite{li2023decoupled} & 80.4  & 85.2  & 87.4  & 84.5  & 75.0  & 71.4  & 75.8  & 75.8  & 72.2  & 63.0  \\
			& LaSA \cite{ji2024language}  & \textbf{80.8} & \textbf{86.7} & \textbf{89.3} & \textbf{86.2} & \textbf{76.3} & \textbf{72.6} & \textbf{79.3} & \textbf{79.3} & \textbf{75.8} & \textbf{66.6} \\
			\midrule
			\multirow{3}{1.7cm}{\centering Point-\\Supervised} 
			& TS-Sup$\,\dag$\cite{li2021temporal} & 59.3  & 63.5  & 63.7  & 57.5  & 39.2  & 55.4  & 50.7  & 51.8  & 46.8  & 33.9  \\
			& TSASPC$\,\dag$\cite{du2022timestamp} &  67.1  & 66.1  & 65.1  & 59.6  & 44.7  & 57.6  & 51.8  & 52.5  & 47.3  & 35.1  \\
			& \textbf{Ours} & \textbf{69.1}  & \textbf{67.5}  & \textbf{67.0}  & \textbf{61.4}  & \textbf{46.5}  & \textbf{59.1}  & \textbf{53.2}  & \textbf{53.8}  & \textbf{48.9}  & \textbf{36.9} \\
			\bottomrule
		\end{tabular}
	}
	\label{tab:MCFS}
\end{table}

\subsection{Main Experimental Results}
Since point-supervised skeleton-based temporal action segmentation is a novel task, there are currently no existing point-supervised methods for direct comparison under the point-supervision setting. We adapt the RGB-video-based point-supervised temporal action segmentation methods of TS-Sup~\cite{li2021temporal} and TSASPC~\cite{du2022timestamp} to operate on skeleton data. We also compare our results with fully supervised skeleton-based temporal action segmentation methods. 

\noindent\textbf{Results on the PKU-MMD: }
Table \ref{tab:PKU} presents experimental results on the PKU-MMD (X-sub) and PKU-MMD (X-view) datasets. Under both evaluation protocols, our method outperforms the two reproduced point-supervised methods on all metrics. Compared with fully supervised methods, our method achieves performance comparable to conventional segmentation methods under the cross-subject evaluation. Under the cross-view evaluation, the Edit and F1@10 metrics even surpass the current state-of-the-art fully supervised method. Due to the lack of precise action boundary information in point supervision, our method falls short of LaSA on stricter F1@25 and F1@50 metrics but still outperforms most fully supervised methods. Additionally, TS-Sup~\cite{li2021temporal} and TSASPC~\cite{du2022timestamp} achieve comparable levels to some advanced fully supervised methods in metrics such as Acc, Edit, and F1@10. This performance gap demonstrates that point supervision can effectively capture the temporal dependencies of actions while reducing the need for frame-level annotations.

\noindent\textbf{Results on the MCFS-22 and MCFS-130: }
Table \ref{tab:MCFS} presents experimental results on the MCFS-22 and MCFS-130 datasets. Under the point-supervision setting, our method consistently outperforms TS-Sup~\cite{li2021temporal} and TSASPC~\cite{du2022timestamp} on both datasets. On the MCFS-22, our method performs slightly worse than traditional fully supervised methods. On the MCFS-130, it achieves comparable performance to traditional methods on Edit, F1@10, and F1@25 metrics, but falls short on Acc and F1@50. This phenomenon may be attributed to the finer-grained category division in MCFS-130, where point supervision exhibits stronger generalization for distinguishing categories, whereas MCFS-22 has fewer categories, making it easier for fully supervised methods to learn the complete temporal structure and thus achieve better performance on this dataset.

\begin{table}[htbp]
	\centering
	\caption{The influence of segmentation models for point-supervised skeleton-based action segmentation on four benchmarks.}
	\resizebox{1\columnwidth}{!}{
		\begin{tabular}{l|ccccc|ccccc|ccccc|ccccc}
			\toprule
			Dataset   & \multicolumn{5}{c|}{PKU-MMD (X-sub)}  & \multicolumn{5}{c|}{PKU-MMD (X-view)} & \multicolumn{5}{c|}{MCFS-22}          & \multicolumn{5}{c}{MCFS-130} \\
			\midrule
			Method   & Acc   & Edit  & F1@10 & F1@25 & F1@50 & Acc   & Edit  & F1@10 & F1@25 & F1@50  & Acc   & Edit  & F1@10 & F1@25 & F1@50 & Acc   & Edit  & F1@10 & F1@25 & F1@50 \\
			\midrule
			MS-TCN~\cite{farha2019ms} & 61.6  & \textbf{67.0}  & \textbf{69.6}  & \textbf{62.5}  & \textbf{44.2}  & \textbf{67.1}  & \textbf{68.8}  & \textbf{73.5}  & \textbf{67.2}  & \textbf{49.5}  & 69.1  & 67.5  & \textbf{67.0}  & \textbf{61.4}  & 46.5  & 59.1  & \textbf{53.2}  & 53.8  & 48.9  & \textbf{36.9}  \\
			MS-TCN++~\cite{li2023ms} & \textbf{61.7}  & 65.6  & 67.8  & 61.1  & \textbf{44.2}  & 66.3  & 67.3  & 71.5  & 65.2  & 48.0 & 69.1  & 67.5  & \textbf{67.0}  & \textbf{61.4}  & 46.5  & 59.1  & \textbf{53.2}  & 53.8  & 48.9  & \textbf{36.9} \\
			\bottomrule
		\end{tabular}
	}
	\label{tab:model-pku}
\end{table}

\begin{table}[h]
	\centering
	
	\begin{minipage}{0.48\textwidth}
		\caption{Ablation studies with pseudo-label integration.}
		\label{tab:321}
		\centering
		 \resizebox{1\columnwidth}{!}{
\begin{tabular}{c|c|c|ccccc}
	\toprule
	Group 1 & Group 2 & Group 3  & Acc & Edit & F1@10 & F1@25 & F1@50 \\
	\toprule
	\ding{51}    &       &       & 60.5  & 64.2  & 65.8  & 58.6  & 40.4  \\
	& \ding{51}     &       & 59.1  & 63.9  & 66.5  & 58.7  & 39.7  \\
	&       & \ding{51}     & 59.7  & 65.0  & 67.1  & 60.3  & 41.2  \\
	\midrule
	\ding{51}     & \ding{51}     &       & 61.3  & 64.6  & 67.7  & 60.8  & 43.5  \\
	\ding{51}     &       & \ding{51}     & 60.3  & 66.0  & 68.8  & 61.4  & 43.5  \\
	& \ding{51}     & \ding{51}     & 60.1  & 64.8  & 67.8  & 61.4  & 43.0  \\
	\midrule
	\ding{51}     & \ding{51}     & \ding{51}     & \textbf{61.6} & \textbf{67.0} & \textbf{69.6} & \textbf{62.5} & \textbf{44.2} \\
	\bottomrule
\end{tabular}}
	\end{minipage}
	\hfill
	\begin{minipage}{0.48\textwidth}
		\caption{Influence of skeleton features on pseudo-label generation methods.}
		\label{tab:xyzF}
		\centering
		 \resizebox{0.9\columnwidth}{!}{
\begin{tabular}{c|c|ccccc}
	\toprule
	Method & Input & Acc   & Edit  & F1@10 & F1@25 & F1@50 \\
	\midrule
	\multirow{3}[2]{*}{Prototype Similarity} & J     & 57.5  & 62.2  & 64.0  & 56.5  & 36.6  \\
	& JF    & 56.9  & 61.2  & 63.2  & 54.8  & 35.7  \\
	& \textbf{J+JF} & \textbf{60.5} & \textbf{64.2} & \textbf{65.8} & \textbf{58.6} & \textbf{40.4} \\
	\midrule
	\multirow{3}[2]{*}{K-Medoids Clustering} & B     & 59.0  & 63.3  & 64.8  & 57.1  & 39.5  \\
	& BF    & 58.0  & 61.2  & 62.6  & 54.2  & 35.6  \\
	& \textbf{B+BF} & \textbf{59.1} & \textbf{63.9} & \textbf{66.5} & \textbf{58.7} & \textbf{39.7} \\
	\midrule
	\multirow{3}[2]{*}{Energy Function} & M     & \textbf{59.7} & 63.4  & 66.6  & 59.5  & 39.9  \\
	& MF    & 58.3  & 62.2  & 64.6  & 57.1  & 39.3  \\
	& \textbf{M+MF} & \textbf{59.7} & \textbf{65.0} & \textbf{67.1} & \textbf{60.3} & \textbf{41.2} \\
	\bottomrule
\end{tabular}}
	\end{minipage}
	
\end{table}

\begin{table}[tbp]
	\centering
	\caption{Influence of multimodal input on pseudo-label generation methods.}
	\resizebox{0.8\columnwidth}{!}{
		\begin{tabular}{ccc|ccccc}
			\toprule
			\makecell{Prototype\\Similarity} & \makecell{K-Medoids\\Clustering} & \makecell{Energy\\Function} & \makecell{Acc} & \makecell{Edit} & \makecell{F1@10} & \makecell{F1@25} & \makecell{F1@50} \\
			\midrule
			J+JF  & J+JF  & J+JF  & 62.2  & 65.7  & 69.0  & 61.6  & 43.7  \\
			B+BF  & B+BF  & B+BF  & \textbf{62.8} & 66.1  & 69.3  & \textbf{62.7} & \textbf{44.7} \\
			M+MF  & M+MF  & M+MF & 61.3  & 64.8  & 67.5  & 61.5  & 42.7  \\
			J+JF  & B+BF  & M+MF  & 61.6  & \textbf{67.0} & \textbf{69.6} & 62.5  & 44.2  \\
			\bottomrule
		\end{tabular}
	}
	\label{tab:duo}
\end{table}


			

\subsection{Ablation Studies and Analysis}

\noindent\textbf{Impact of action segmentation model: }
Table \ref {tab:model-pku} presents the performance of different action segmentation networks, such as MS-TCN~\cite {farha2019ms} and MS-TCN++~\cite {li2023ms}. The results demonstrate that the differences between MS-TCN and MS-TCN++ across various evaluation metrics are relatively minor. Although the segmentation model contributes to some extent in our method, it is not a decisive factor. 

\noindent\textbf{Ablation studies with pseudo-label integration: }
Table \ref{tab:321} presents results of different pseudo-label generation methods. Among them, `Group 1' represents a prototype similarity method that uses joint data and joint features as input; `Group 2' represents the constrained K-Medoids clustering that uses bone data and bone features as input; `Group 3' represents the energy function that uses motion data and motion characteristics as input. It can be observed from the experimental results that the overall performance when using two pseudo-label generation methods simultaneously is better than that when using only one method. When the three pseudo-label generation methods are used in an integrated way, the model achieves the optimal effect, which verifies the effectiveness and necessity of pseudo-label integration.


\noindent\textbf{Impact of modal selection for pseudo-label generation: }
According to Table \ref{tab:xyzF}, when using the prototype similarity method to generate pseudo-labels, the combination of joint data (J) and joint features (JF) yields better performance than using either one alone, while the original joint data generally outperforms the joint features. A similar trend is observed when using the constrained K-Medoids clustering to generate pseudo-labels for bone data (B) and bone features (BF): the combination of data and features leads to improved results, and bone data shows superior performance compared to bone features. Likewise, for motion data (M) and motion features (MF), where an energy function is used to generate pseudo-labels, the results align with those of joint and bone data—the integration of original data and features enhances performance, and motion data performs better than motion features. Across different modalities, the fusion of original skeleton and features contributes to improved pseudo-label quality, with original skeleton playing a more significant role in enhancing the generation outcome.

To investigate the impact of different pseudo-label generation inputs on action segmentation, we standardize the inputs for the three pseudo-label generation methods. Specifically, all methods utilize three input modalities: joint data (J) and joint features (JF), bone data (B) and bone features (BF), and motion data (M) and motion features (MF). Experimental results in Table \ref{tab:duo} indicate that when using a single modality as input, the combination of bone data and bone features performs the best, achieving the highest scores across evaluation metrics such as Acc, F1@25, and F1@50, whereas the combination of motion data and motion features yields relatively poorer results. When the three modalities are fused, the experimental results demonstrate improved stability. This is because multimodal fusion provides richer information from diverse perspectives, thus enhancing the robustness of the outcomes. 

\begin{figure*}[tbp]
	\centering
	\includegraphics[width=1\textwidth]{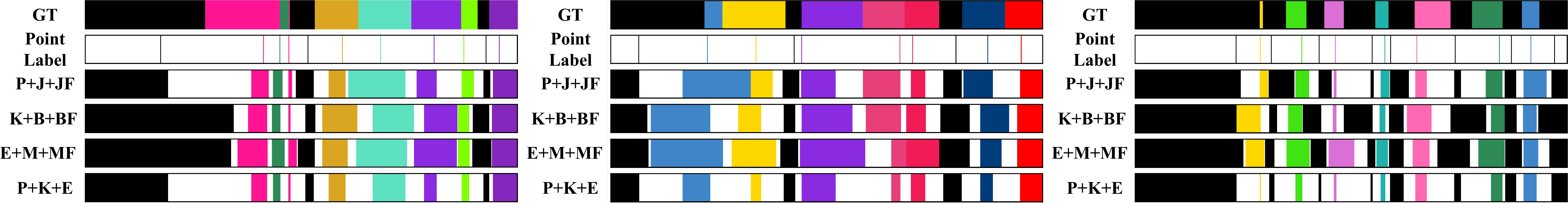} 
	\caption{Comparison of results of different pseudo-label generation methods on the PKU-MMD (X-sub) dataset.}
	\label{fig:pseudo}
\end{figure*}

\begin{figure*}[tbp]
	\centering
	\includegraphics[width=1\textwidth]{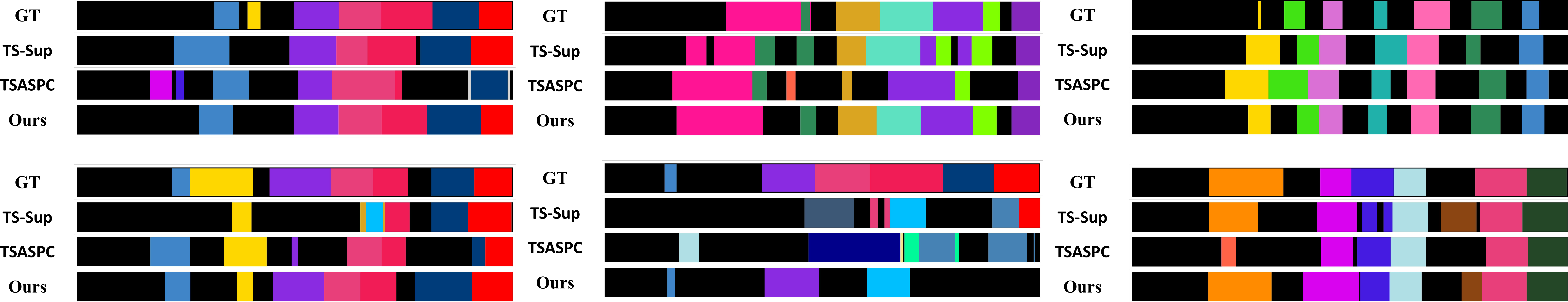} 
	\caption{Comparison of action segmentation results of different point-supervised methods on the PKU-MMD (X-sub) dataset.}
	\label{fig:seg_res}
\end{figure*}


\noindent\textbf{Visualizations of generated pseudo labels: }
Fig. \ref{fig:pseudo} visualizes the label outputs generated by multiple pseudo-labeling methods on the PKU-MMD (X-sub) dataset. Each example consists of six label visualizations, with distinct colors representing different action categories: The first row (GT) presents the ground truth labels under full annotation; the second row presents sparsely annotated point-wise labels; the third row (P+J+JF) presents pseudo-labels generated via the prototype similarity method (P) using joint data (J) and joint features (JF); the fourth row (K+B+BF) presents pseudo-labels produced by the constrained K-Medoids clustering method using bone data (B) and bone features (BF); the fifth row (E+M+MF) presents pseudo-labels created using an energy function with motion data (M) and motion features (MF); the sixth row (P+K+M) presents the integrated pseudo-labels, combining the three methods.

From the second to the sixth row, blank areas represent ambiguous intervals where no specific action category is assigned to the corresponding frames. It can be observed that the integrated method generates pseudo-labels with larger blank regions compared to a single method. This is because the integrated mechanism eliminates frames where there are differences among algorithms. Furthermore, since all the above methods rely on the original skeleton data and its features for label generation, there is no category assignment for the frames with differences among algorithms when generating pseudo-labels. This also leads to a certain proportion of blank areas in the labels generated by each method.

\noindent\textbf{Visualizations of temporal action segmentation: }
Fig. \ref{fig:seg_res} visualizes the prediction results of different point-supervised methods on the PKU-MMD (X-sub) dataset. Visualized results for each video are presented in four rows, with each color representing a distinct action category: The first row presents the ground truth labels; the second row presents the predictions of TS-Sup~\cite{li2021temporal}; the third row presents the results of TSASPC~\cite{du2022timestamp}; and the fourth row presents the predictions of the proposed method. Comparative visualization results demonstrate that the action predictions of our method are closer to the ground truth labels than those of the other two methods.

\section{Conclusion}

In this paper, we present a novel task of point-supervised skeleton-based temporal action segmentation. Three modalities of skeleton data, i.e., joint, bone, and motion, are processed by the pretrained unified model to extract high-dimensional feature representations, providing multimodal inputs for the subsequent pseudo-label generation and integration method. For pseudo-label generation, we introduce a prototype similarity method and combine it with two existing pseudo-label generation methods to produce diverse pseudo-labels. By integrating pseudo-labels generated from different input modalities, the robustness and accuracy of the pseudo-labels are further enhanced, ensuring that the generated pseudo-labels can effectively guide model training. We adopt the action segmentation model for training, supervised by the generated pseudo-labels. Extensive experiments on four commonly used benchmarks demonstrate that our method achieves outstanding performance across multiple evaluation metrics, even surpassing state-of-the-art fully supervised methods in some metrics on certain datasets. Our method validates the effectiveness of point supervision in skeleton-based temporal action segmentation. By introducing point supervision, not only is annotation time significantly reduced, but semantic ambiguities in action boundaries are also effectively mitigated.

\vspace{0.3cm}
\noindent \textbf{Acknowledgements.} This research was supported by the National Natural Science Foundation of China (Nos. 62302093 and 52441503) and the Natural Science Foundation of Jiangsu Province (No. BK20230833).

\vspace{0.1cm}
\noindent \textbf{Disclosure of Interests.} The authors have no competing interests to declare that are relevant to the content of this article.

%
%
%
\bibliographystyle{splncs04}
\bibliography{IEEEfull}

\end{document}